\documentclass[sigconf]{acmart}

\usepackage{multirow}
\usepackage{tabularx}
\usepackage{subcaption}

\AtBeginDocument{%
 }

\setcopyright{acmcopyright}

\copyrightyear{2023}
\acmYear{2023}
\setcopyright{rightsretained}
\acmConference[SIGIR '23]{Proceedings of the 46th International ACM SIGIR Conference on Research and Development in Information Retrieval}{July 23--27, 2023}{Taipei, Taiwan}
\acmBooktitle{Proceedings of the 46th International ACM SIGIR Conference on Research and Development in Information Retrieval (SIGIR '23), July 23--27, 2023, Taipei, Taiwan}\acmDOI{10.1145/3539618.3592069}
\acmISBN{978-1-4503-9408-6/23/07}

\makeatletter
\gdef\@copyrightpermission{
  \begin{minipage}{0.3\columnwidth}
   \href{https://creativecommons.org/licenses/by/4.0/}{\includegraphics[width=0.90\textwidth]{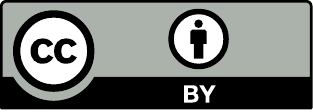}}
  \end{minipage}\hfill
  \begin{minipage}{0.7\columnwidth}
   \href{https://creativecommons.org/licenses/by/4.0/}{This work is licensed under a Creative Commons Attribution International 4.0 License.}
  \end{minipage}
  \vspace{5pt}
}
\makeatother

\begin{document}

\title[Text-to-Motion Retrieval: Towards Joint Understanding of Human Motion Data and Natural Language]{Text-to-Motion Retrieval: Towards Joint Understanding of \\Human Motion Data and Natural Language}




\author{Nicola Messina}
\authornote{Both authors contributed equally to the paper.}
\affiliation{%
  \institution{ISTI-CNR}
  \city{Pisa}
  \country{Italy}}
\email{nicola.messina@isti.cnr.it}

\author{Jan Sedmidubsky}
\authornotemark[1]
\affiliation{%
  \institution{Masaryk University}
  \city{Brno}
  \country{Czechia}}
\email{xsedmid@fi.muni.cz}

\author{Fabrizio Falchi}
\affiliation{%
  \institution{ISTI-CNR}
  \city{Pisa}
  \country{Italy}}
\email{fabrizio.falchi@isti.cnr.it}

\author{Tom\'{a}\v{s} Rebok}
\affiliation{%
  \institution{Masaryk University}
  \city{Brno}
  \country{Czechia}}
\email{rebok@ics.muni.cz}

\renewcommand{\shortauthors}{Nicola Messina, Jan Sedmidubsky, Fabrizio Falchi, \& Tom\'{a}\v{s} Rebok}
\newcommand\ourarchitecture{MoT}

\begin{abstract}
Due to recent advances in pose-estimation methods, human motion can be extracted from a common video in the form of 3D skeleton sequences. Despite wonderful application opportunities, effective and efficient content-based access to large volumes of such spatio-temporal skeleton data still remains a challenging problem. In this paper, we propose a novel content-based text-to-motion retrieval task, which aims at retrieving relevant motions based on a specified natural-language textual description. To define baselines for this uncharted task, we employ the BERT and CLIP language representations to encode the text modality and successful spatio-temporal models to encode the motion modality. We additionally introduce our transformer-based approach, called Motion Transformer (\ourarchitecture{}), which employs divided space-time attention to effectively aggregate the different skeleton joints in space and time. Inspired by the recent progress in text-to-image/video matching, we experiment with two widely-adopted metric-learning loss functions. Finally, we set up a common evaluation protocol by defining qualitative metrics for assessing the quality of the retrieved motions, targeting the two recently-introduced KIT Motion-Language and HumanML3D datasets. The code for reproducing our results is available here: \url{https://github.com/mesnico/text-to-motion-retrieval}.
\end{abstract}

\ccsdesc[500]{Information systems~Novelty in information retrieval}
\ccsdesc[300]{Information systems~Language models}
\ccsdesc[300]{Information systems~Evaluation of retrieval results}
\ccsdesc[300]{Information systems~Data access methods}

\keywords{human motion data, skeleton sequences, CLIP, BERT, deep language models, ViViT, motion retrieval, cross-modal retrieval}
\begin{teaserfigure}
  \centering
  \includegraphics[width=0.48\textwidth]{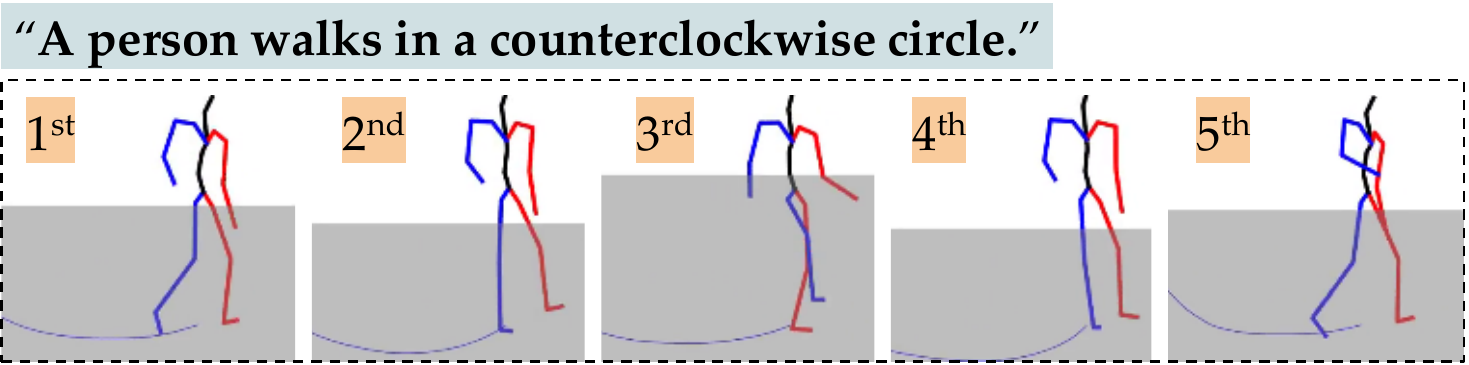}
  \quad
  \includegraphics[width=0.48\textwidth]{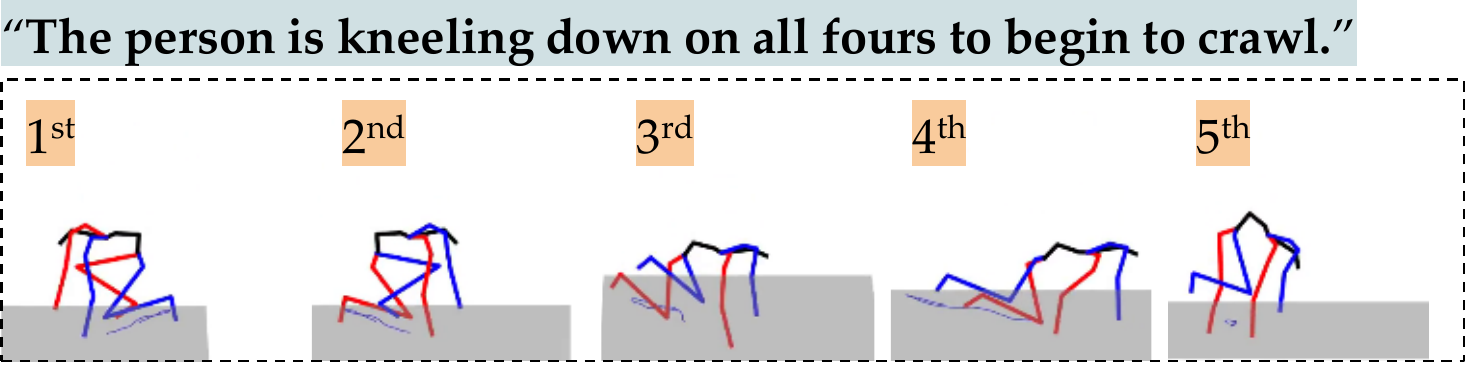}
  \caption{Five motions retrieved for two different queries specified by free text (CLIP as text encoder, \ourarchitecture\ as motion encoder).}
  \label{fig:teaser}
\end{teaserfigure}


\maketitle

\section{Introduction and Related Work}

Pose-estimation methods~\cite{DD22} can detect 3D human-body keypoints in a single RGB video stream.
The keypoints detected in individual frames constitute a simplified spatio-temporal representation of human motion in the form of a so-called \emph{skeleton sequence}.
As indicated in~\cite{SEBZ21-survey}, the analysis of such representation opens unprecedented application potential in many domains, ranging from virtual reality, through robotics and security, to sports and medicine.
%
%
The ever-increasing popularity of skeleton data calls for technologies able to effectively and efficiently access large volumes of such spatio-temporal data based on its content.

%
Research in skeleton-data processing mainly focuses on designing deep-learning architectures for classification of labeled actions \cite{CCCWZL21,PHZ21,LWFP21} or detection of such actions in continuous streams~\cite{PGBA19,SLXZL18}. The proposed architectures are often learned in a supervised way based on transformers~\cite{AKC21,CCCWZL21,CCZL20}, convolutional~\cite{LWFP21}, recurrent~\cite{SLXZL18}, or graph-convolutional~\cite{DWCL22,PHZ21} networks. Recently, self-supervised methods are becoming increasingly popular as they can learn motion semantics without knowledge of labels using reconstruction-based~\cite{ZZCHZZLS23,SCA23-ecir} or contrastive-based learning~\cite{YLG22,LSYL20}.

The trained architectures can serve as \emph{motion encoders} that express the motion semantics by a high-dimensional \emph{feature} vector extracted from the last hidden network layer. This concept can be transferred to the \emph{motion retrieval} task to support content-based access based on the \emph{query-by-example} paradigm~\cite{SEBZ21-survey,BSZ21-icmr,SBDZ20-ecir}, which aims at identifying the database motions that are the most similar to a user-defined query motion. Besides balancing descriptiveness and indexability of the motion features, the most critical issue is to specify a convenient query motion example. The example can be selected from available skeleton sequences~\cite{SCA23-ecir}, drawn in a visualization-driven graphical user interface~\cite{BWKMSK13}, modeled by puppet interfaces~\cite{NNSH11}, specified as a set of logical constraints~\cite{KCTBK13}, or artificially generated~\cite{DGL09}. However, such a query example may not ever exist, or its construction requires professional modeling skills.
This paper focuses on motion retrieval but simplifies query specification by enabling users to formulate a query by free text.

With the current advances in cross-modal learning, especially in the field of textual-visual processing, the trend is to learn common multi-modal spaces~\cite{messina2022transformer} so that similar images can be described and searched with textual descriptions~\cite{messina2021towards}. A representative example is the CLIP model~\cite{RKHRGASAMCKS21-CLIP}, which learns an effective common space for the visual and textual modalities. This allows the use of open vocabularies or complex textual queries for searching images.

Our work has many analogies with the text-to-video retrieval task \cite{fang2021clip2video,luo2022clip4clip,shvetsova2022everything,zhao2022centerclip,liu2022animating}, given that the moving skeleton also evolves in space and time.
Despite the popularity of such powerful and versatile text-vision models, no effort has been made for the skeleton-data modality. Differently from video data, the skeleton is anonymized and avoids learning many common biases present in video datasets. To the best of our knowledge, there is only one approach~\cite{kim2022learning} that relates to text-to-motion matching. However, it uses pre-training and tackles only the classification task. A few available datasets providing the training data for text-to-motion retrieval -- e.g., the KIT Motion Language~\cite{Plappert2016-KITdataset} and recently-released HumanML3D~\cite{GZZWJLC22-HumanML3D} datasets -- are primarily used for motion generation from a textual description~\cite{ZZCHZZLS23,GZWC22,TRGSCB22,ZCPHGYL22,PBV21}, where the idea is to align text and motion embeddings into a common space, but never explicitly handling the \textit{text-to-motion} retrieval task.

\subsection*{Contributions of this Paper}
We tackle the above-mentioned gap by introducing a novel \emph{text-to-motion} retrieval task, which aims at searching databases of skeleton sequences and retrieving those that are the most relevant to a detailed textual query. 
For this task, we define evaluation metrics, establish new qualitative baselines, and propose the first text-to-motion retrieval approach. 
These initial contributions can be employed for future studies on this challenging yet unexplored task.

Specifically, one of the main paper contributions is the proposal of a fair baseline by adopting promising (1) \emph{motion encoders} already employed as backbones in other motion-related tasks and (2) \emph{text encoders} successfully applied in natural language processing (NLP) and text-to-image retrieval. The core of this baseline is a two-stream pipeline where the motion and text modalities are processed by separate encoders. The obtained representations are then projected into the same common space, for which a metric is learned in a similar way as in CLIP~\cite{RKHRGASAMCKS21-CLIP} or ALADIN~\cite{messina2022aladin} in the text-to-image scenario.
The choice of a two-stream pipeline is strategic to make the approach scalable to large motion collections, as feature vectors extracted from both modalities can be easily stored in off-the-shelf indexes implementing efficient similarity search access.

Inspired by recent advances in video processing~\cite{ADHSLS21}, we also propose a transformer-based motion encoder -- the Motion Transformer (\ourarchitecture) -- that employs divided space-time attention on skeleton joints. We show that \ourarchitecture{} reaches competitive results with respect to a state-of-the-art motion encoder, DG-STGCN \cite{DWCL22}, on both KIT Motion Language and HumanML3D datasets.





\begin{figure}[t]
  \centering
  \includegraphics[width=\linewidth]{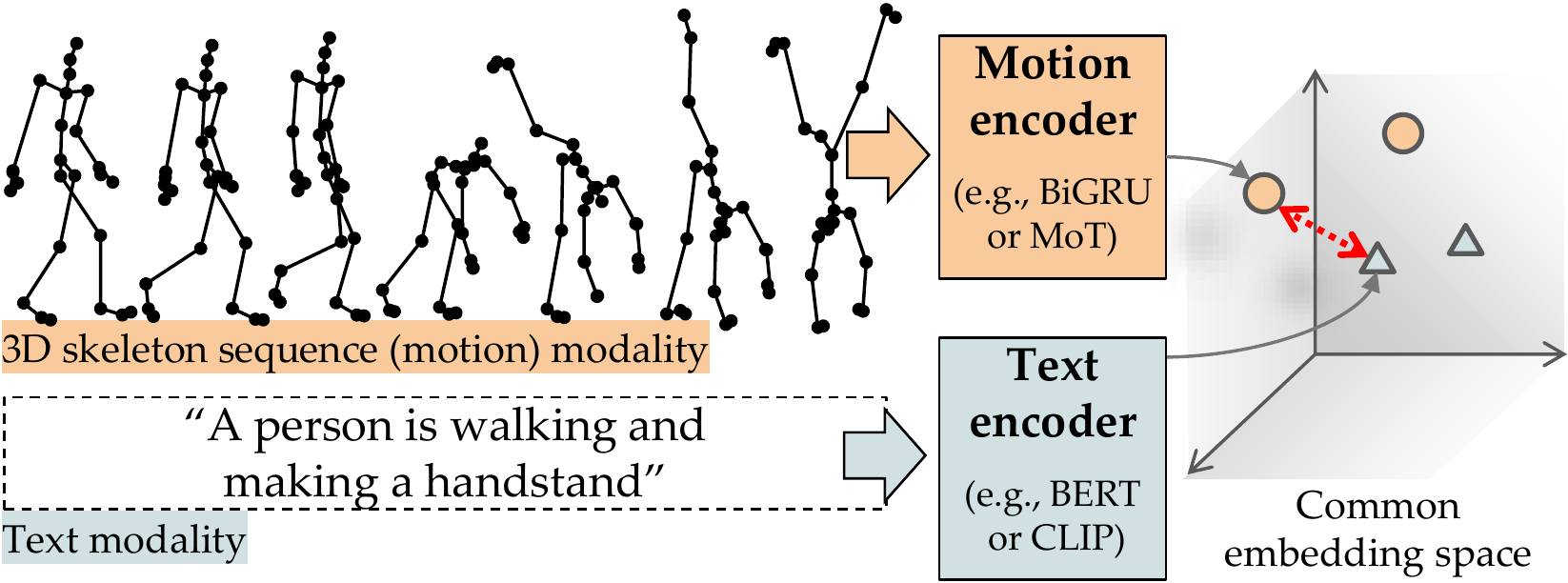}
  \caption{Schematic illustration of the learning process of the common space of both the text and motion modalities.}
  \label{fig:schema}
\end{figure}

\section{Text-to-Motion Retrieval Pipeline}

The main idea of our approach is to rely on a two-stream pipeline, where motion and text features are first extracted through ad-hoc encoders and then projected into the same common space, as schematically illustrated in Figure~\ref{fig:schema}.
In this section, we sketch the whole pipeline which consists of the: (i) text encoder, (ii) motion encoder, and (iii) loss function used to optimize the common space.

\subsection{Text Encoders}
Inspired by recent works in NLP, we rely on two pre-trained textual models, namely BERT~\cite{kenton2019bert} and the textual encoder from CLIP~\cite{RKHRGASAMCKS21-CLIP}.

\textbf{BERT}.
We use the implementation from \cite{ghosh2021synthesis}, which performed the task of motion synthesis conditioned on a natural language prompt. This model stacks together a BERT pre-trained module and an LSTM model composed of two layers for aggregating the BERT output tokens, producing the final text embedding. 
We take the final hidden state of the LSTM model as our final sentence representation.  
As in \cite{ghosh2021synthesis}, the BERT model is fixed. At training time, we only update the LSTM weights.

\textbf{CLIP}. It is a recently-introduced vision-language model trained in a contrastive manner for projecting images and natural language descriptions in the same common space~\cite{RKHRGASAMCKS21-CLIP}. 
Here, we use the textual encoder of CLIP, which is composed of a transformer encoder \cite{vaswani2017attention} with modifications introduced in \cite{radford2019language}, and employs lower-cased byte pair encoding (BPE) representation of the text. 
We then stack an affine projection to the CLIP representation, which -- similarly to the BERT+LSTM case -- is the only layer to be trained.

\subsection{Motion Encoders}
Differently from the textual pipeline, which takes as input an unstructured natural language sentence, the input to motion encoder models is a vector $\mathbf{x}\in\mathbb{R}^{T\times J\times D}$, where $T$ is the time length of the motion, $J$ is the number of joints of the human-body skeleton, and $D$ is the number of features used to encode each joint.

\textbf{Bidirectional GRU}. This architecture is widely adopted in time-series processing, and an early variant that used LSTM was applied to frame-level action detection in continuous motion data \cite{CESZ19-mtap}.
In particular, we first increase the dimensionality of the input -- which is $D=9$ in our case -- by using a two-layer feed-forward network (FFN) before feeding it into the GRU: $\overrightarrow{\bar{\mathbf{z}}}, \overleftarrow{\bar{\mathbf{z}}} = \overleftrightarrow{\text{GRU}}(\text{FFN}(\mathbf{x}))$. Then, we compute the final motion embedding by concatenating the representations $\overrightarrow{z}$ and $\overleftarrow{z}$. 

\textbf{Upper-Lower GRU}. To better learn semantics of different body parts, we adopt the model in \cite{ghosh2021synthesis} to independently process the upper and lower parts of the skeleton using two GRU layers.

\textbf{DG-STGCN}. This architecture~\cite{DWCL22} recently reached state-of-the-art results in motion classification. Their GCN module features a spatial module, built of affinity matrices to capture dynamic graphical structures, and a temporal module that performs temporal aggregation using group-wise temporal convolutions. We refer the reader to the original formulation \cite{DWCL22} for further details.

\textbf{\ourarchitecture}. Our proposed architecture that we built on top of the successful transformer-based video processing network ViViT~\cite{ADHSLS21}. In the original implementation, which processes a sequence of frames, the dimension $J$ is the number of grid-arranged rectangular patches from each frame. In our case, instead, the spatial features come from the joints. Instead of using as $J$ all individual skeleton joints, we first aggregate them obtaining features for five different body parts, similar to the pre-processing performed in Upper-Lower GRU. In this way, $J=5$, which is far less than the total number of skeleton joints. This is beneficial from a computational point of view, and we found that this solution also reaches the best performance.

\subsection{Optimization}
We explore two widely-adopted metric learning loss functions, namely the symmetric triplet loss widely used in text-to-image \cite{messina2021fine} and the 
InfoNCE Loss, introduced for cross-modal matching in \cite{zhang2020contrastive} and employed in CLIP~\cite{RKHRGASAMCKS21-CLIP} and recent cross-modal works \cite{luo2022clip4clip}. We assume $(\mathbf{m}_i, \mathbf{c}_i)$ is the $i$-th motion and caption embedding pair, $S(\cdot, \cdot)$ is the cosine similarity, and $B$ is the batch size.

The symmetric triplet loss is defined as:
\begin{equation*}
\frac{1}{B} \sum_i^B \max_{j, j\neq i} [\alpha + S(\mathbf{m}_i,\mathbf{c}_j) - S(\mathbf{m}_i,\mathbf{c}_i)]_+ + 
\max_{j, j\neq i} [\alpha + S(\mathbf{m}_j,\mathbf{c}_i) - S(\mathbf{m}_i,\mathbf{c}_i)]_+
\end{equation*}
where $[x]_+ \equiv max(0, x)$ and $\alpha$ is a fixed margin. The index $j$ identifies the hardest negative of the element with index $i$.

Info-NCE is basically a symmetric cross-entropy loss, defined as:
\begin{equation*}
\begin{split}
    - \frac{1}{B} \sum_i^B \log \frac{\exp (S(\mathbf{m}_i,\mathbf{c}_i)/\tau)}{\sum_j^B \exp(S(\mathbf{m}_i,\mathbf{c}_j)/\tau)} + \log \frac{\exp (S(\mathbf{m}_i,\mathbf{c}_i)/\tau)}{\sum_j^B \exp(S(\mathbf{m}_j,\mathbf{c}_i)/\tau)}
\end{split}
\end{equation*}
where $\tau$ is a temperature parameter learned during training.

\section{Experimental Evaluation}

\begin{table*}
\tabcolsep 4pt
\renewcommand{\arraystretch}{0.80}
\caption{Text-to-motion retrieval results on both the KIT Motion Language Dataset and HumanML3D Dataset. We report the best and the second-best results with bold and underlined font, respectively.}
\label{tab:results}
\begin{tabular}{ll|ccccccc|ccccccc}
\toprule
& &
\multicolumn{7}{c|}{\textbf{KIT Motion Language Dataset}} & \multicolumn{7}{c}{\textbf{HumanML3D Dataset}} \\
& &
\multicolumn{3}{c}{Recall@k $\uparrow$} & \multicolumn{2}{c}{Rank $\downarrow$} & \multicolumn{2}{c|}{nDCG $\uparrow$} &
\multicolumn{3}{c}{Recall@k $\uparrow$} & \multicolumn{2}{c}{Rank $\downarrow$} & \multicolumn{2}{c}{nDCG $\uparrow$} \\
\cmidrule(lr){3-5} \cmidrule(lr){6-7} \cmidrule(lr){8-9}
\cmidrule(lr){10-12} \cmidrule(lr){13-14} \cmidrule(lr){15-16}
\textbf{Text Model} & \textbf{Motion Model} & r1 & r5 & r10 & mean & med & \texttt{SPICE} & \texttt{spaCy}
& r1 & r5 & r10 & mean & med & \texttt{SPICE} & \texttt{spaCy} \\
\midrule
\multirow[c]{4}{*}{BERT+LSTM} & BiGRU & 3.7 & 15.2 & 23.8 & 72.3 & 30 & 0.271 & 0.706
 & 2.9 & 11.8 & 19.8 & 253.9 & 55 & 0.250 & 0.768 \\
 & UpperLowerGRU & 3.2 & 15.7 & 25.3 & 90.2 & 34 & 0.263 & 0.697
 & 2.4 & 10.5 & 17.7 & 285.7 & 68 & 0.242 & 0.763 \\
 & DG-STGCN & 6.2 & 24.5 & \underline{38.2} & 40.6 & 17 & 0.339 & 0.740
 & 2.0 & 8.4 & 14.4 & 242.0 & 73 & 0.231 & 0.767 \\
 & \ourarchitecture & 5.3 & 21.3 & 32.0 & 51.1 & 20 & 0.318 & 0.723
 & 2.5 & 11.2 & 19.4 & 234.5 & 51 & 0.247 & 0.768 \\
\cline{1-16}
\multirow[c]{4}{*}{CLIP} & BiGRU & \underline{6.6} & 21.5 & 32.3 & 52.0 & 22 & 0.316 & 0.729
 & 3.4 & 14.3 & 23.1 & 201.9 & 43 & 0.272 & 0.780 \\
 & UpperLowerGRU & 6.4 & 22.0 & 32.2 & 52.3 & 22 & 0.321 & 0.732
 & 3.1 & 12.6 & 20.8 & 200.4 & 47 & 0.269 & 0.779 \\
 & DG-STGCN & \textbf{7.2} & \underline{26.1} & \underline{38.2} & \underline{36.9} & \underline{16} & \textbf{0.355} & \textbf{0.751}
 & \textbf{4.1} & \textbf{16.0} & \textbf{26.5} & \textbf{159.6} & \textbf{33} & \textbf{0.291} & \textbf{0.789} \\
 & \ourarchitecture & 6.5 & \textbf{26.4} & \textbf{42.6} & \textbf{35.5} & \textbf{14} & \underline{0.352} & \underline{0.748}
 & \underline{3.5} & \underline{14.8} & \underline{24.5} & \underline{166.2} & \underline{38} & \underline{0.280} & \underline{0.785} \\
\bottomrule
\end{tabular}
\end{table*}

\subsection{Metrics}


\paragraph{Exact-search}
Exact-search metrics leverage the intrinsic ground truth available in the employed datasets, where motions come with one (or more) textual descriptions. We can consider motions associated with the given textual query as the \textit{exact} solutions, while all the other ones as irrelevant by default. In this context, the \textbf{recall@k} measures the percentage of queries that find the correct result within the first \textit{k} elements in the results list, while the median and mean ranks 
represent the median and mean rank of the exact result computed among all the queries.

\paragraph{Relevance-based}
There can exist motions relevant to a certain extent to the given textual query that are not paired in the dataset. In this context, the normalized Discounted Cumulative Gain (nDCG) metric is widely employed. 
The DCG takes into consideration the \textit{relevance} a specific item has with the query, discounting it with a logarithmic factor that depends on the rank of that item:  $\text{DCG}_p = \sum_{i=1}^p \frac{2^{\text{rel}_i} - 1}{\log_2{(i+1)}}.$
%
%
The nDCG normalizes DCG by its maximum theoretical value and thus returns values in the $[0,1]$ range.
We define the relevance similarly to previous works in image-to-text retrieval \cite{messina2021fine,messina2021transformer,carrara2018picture}, that use a proxy relevance between textual descriptions, which is much easier to compute. In this work, we use two textual relevance functions: (i) the \texttt{SPICE} relevance~\cite{AndersonFJG16spice} -- a hand-crafted relevance that exploits graphs associated with the syntactic parse trees of the sentences and has a certain degree of robustness against synonyms; and (ii) the \texttt{spaCy} relevance obtained from the \textit{spaCy} Python tool, which implements a deep learning-powered similarity score for pairs of texts.

\subsection{Datasets and Evaluation Protocol}
We employ two recently introduced datasets, HumanML3D~\cite{GZZWJLC22-HumanML3D} and KIT Motion Language~\cite{Plappert2016-KITdataset}. Both datasets carry one or more human-written descriptions for each motion. We employ the same pre-processing pipeline for both datasets -- the one developed in the codebase of the HumanML3D dataset~\cite{GZZWJLC22-HumanML3D}. 
We employ $D=9$ features to represent each joint: six features encoding continuous rotation representation plus three features encoding rotation-invariant forward kinematics joint positions.

KIT Motion-Language Dataset contains 3,911 recordings of full-body motion in the Master Motor Map form~\cite{terlemez2014master}, along with textual descriptions for each motion. It has a total of 6,278 annotations in English, where each motion recording has one or more annotations that explain the action, like \textit{"A human walks two steps forwards, pivots 180 degrees, and walks two steps back".} 

HumanML3D is, in its essence, very similar to KIT Motion Language Dataset. However, it is a more recent dataset developed by adding textual annotations to already-existing and widely-used motion-capture datasets -- AMASS~\cite{mahmood2019amass} and HumanAct12~\cite{guo2020action2motion}. It contains 14,616 motions annotated by 44,970 textual descriptions. 

The results are reported on the test set of the respective datasets after removing possibly redundant queries. In particular, we use 938 and 8,401 textual queries to search among 734 and 4,198 motions for the KIT and HumanML3D datasets, respectively. For HumanML3D, these motions are obtained by splitting the originally provided ones using the available segment annotations associating a motion subsequence with the text that describes it. In this sense, HumanML3D enables a finer retrieval, as texts are more likely to describe the correct subsequence instead of the whole motion.





  

\subsection{Results}
We report text-to-motion retrieval results in Table~\ref{tab:results}, obtained with the InfoNCE loss (see Section \ref{sec:ablation} for a comparison of loss functions).
The best results are competitively achieved by both DG-STGCN and our transformer-based \ourarchitecture.
The first remarkable insight is the superiority of CLIP over the BERT+LSTM on all the metrics in both datasets. With CLIP, the effectiveness of DG-STGCN and \ourarchitecture\ over GRU-based methods is evident, especially on the KIT dataset, where the mean rank is almost 30\,\% lower. The nDCG metric, through the highly-semantic text-based relevance scores, confirms the trend of the recall@k values, suggesting that the CLIP model paired with GCNs and Transformers can both retrieve exact and relevant results in earlier positions in the results list. Notably, from an absolute perspective, all the methods reach overall low performance on exact search, confirming the difficulty of the introduced text-to-motion retrieval task. This may be due to (i) some intrinsic limitations that are hard to eliminate -- e.g., textual descriptions are written by annotators by possibly looking at the original video, which the network has no access to -- or (ii) difficulties in capturing high-level semantics in motion or text data. 
\begin{figure}
 \centering
 \begin{subfigure}[t]{.49\linewidth}
     \centering
     \includegraphics[trim=.5cm .5cm .5cm 0cm, clip, width=\linewidth]{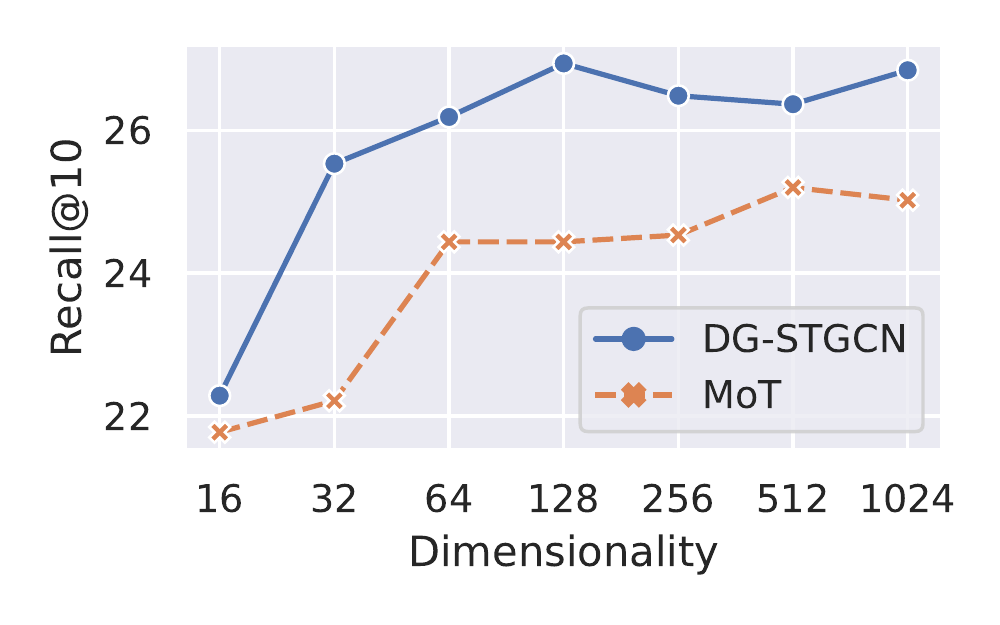}
     \caption{Recall@10}
     \label{fig:r10-vs-dim}
 \end{subfigure}
 \begin{subfigure}[t]{.49\linewidth}
    \centering
    \includegraphics[trim=.5cm .5cm .5cm 0cm, clip, width=\linewidth]{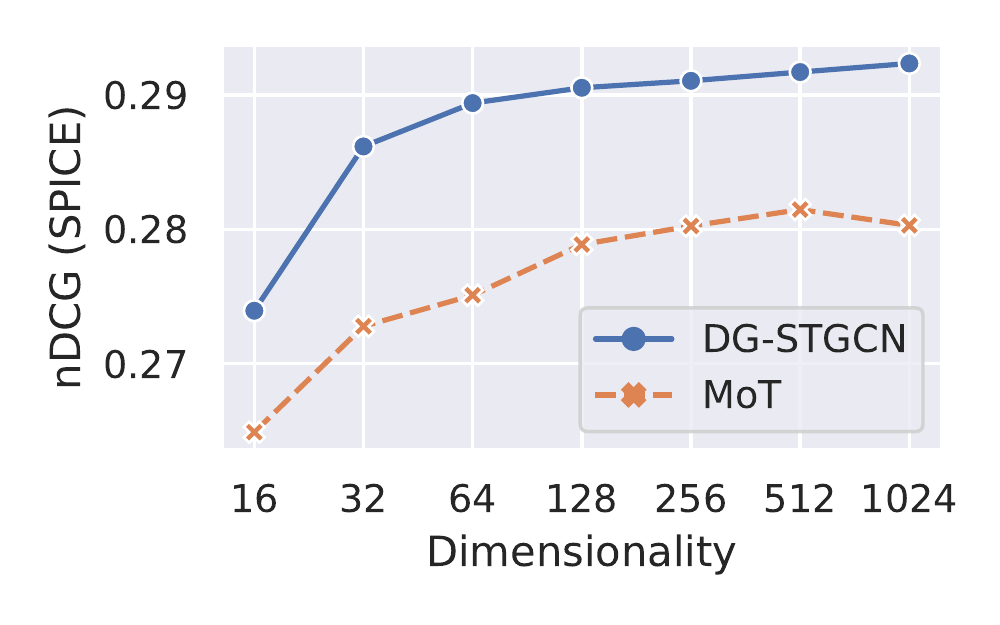}
    \caption{nDCG (\texttt{SPICE})}
    \label{fig:spice-vs-dim}
 \end{subfigure}
 \caption{Performance varying space dimensionality.}
 \label{fig:sweep-dim}
 \end{figure}
\begin{figure}
 \centering
 \begin{subfigure}[t]{.49\linewidth}
  \centering
  \includegraphics[trim=.5cm .5cm .5cm .5cm, clip, width=\linewidth]{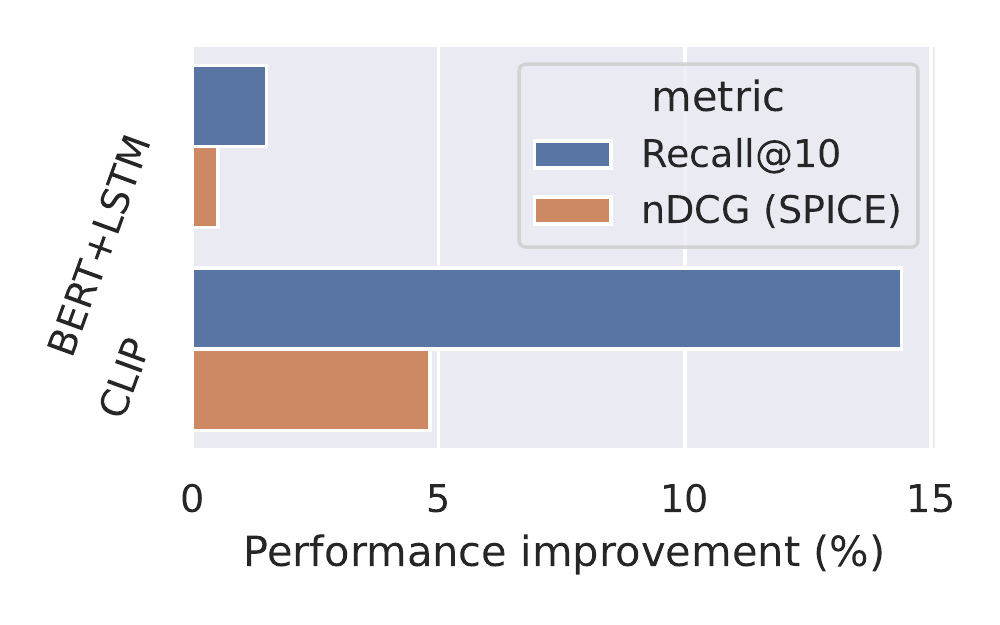}
  \caption{Text models dependence}
  \label{fig:nce-vs-triplet-text}
 \end{subfigure}
 \begin{subfigure}[t]{.49\linewidth}
 \centering
  \includegraphics[trim=.5cm .5cm .5cm .5cm, clip, width=\linewidth]{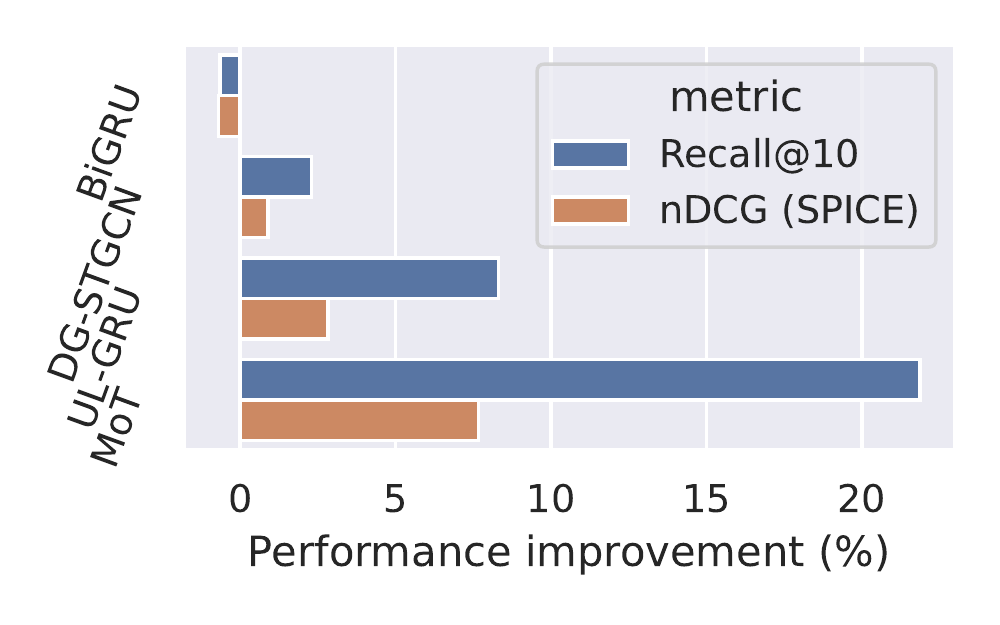}
  \caption{Motion models dependence}
  \label{fig:nce-vs-triplet-motion}
 \end{subfigure}
\caption{Improvement of InfoNCE loss over Triplet loss.}
\label{fig:nce-vs-triplet}
\end{figure}
%
%
In Figure~\ref{fig:teaser}, we report two qualitative examples of text-to-motion retrieval using CLIP + \ourarchitecture, on HumanML3D. We can notice the potential of such natural-language-based approach to motion retrieval. Specifically, note how the approach is sensible to asymmetries -- in the first case, where the \textit{counterclockwise} adjective is specified in the query, only the correctly-oriented motions are returned in the first positions; in the second case, where no \textit{right} or \textit{left} is specified, both the original and mirrored motions are returned (e.g., the 1st and 2nd results).

\subsubsection{Ablation Study on Loss Function and Space Dimensionality}
\label{sec:ablation}
In Figure~\ref{fig:sweep-dim}, we report performance when varying the dimensionality of the common space, for the two motion models DG-STGCN and \ourarchitecture\ employing the CLIP text model. We can notice how, on both metrics in Figure \ref{fig:r10-vs-dim}/\ref{fig:spice-vs-dim}, the effectiveness remains quite high even for very small dimensions of the common space, with a negligible improvement after 256 dimensions. Specifically, with only 16 dimensions instead of 256, the performance drops by only about 6\,\% on nDCG with \texttt{SPICE} relevance and on average 15\,\% on Recall@10, considering both motion encoders. This suggests that the intrinsic dimensionality of the learned space is quite small, opening the way for further studies and feature visualization in future works.

In Figure~\ref{fig:nce-vs-triplet}, we also report the remarkable performance gain achieved by InfoNCE loss over the standard symmetric triplet loss. We can see how the InfoNCE loss induces the best results basically in all the configurations, confirming its power even in the under-explored text-motion joint domain. Breaking down the contributions of this variation on the text and motion models in Figures \ref{fig:nce-vs-triplet-text} and \ref{fig:nce-vs-triplet-motion} respectively, we notice how the best gains are achieved by using the CLIP textual model and the \ourarchitecture\ motion model.

\section{Conclusions}
In this paper, we introduced the task of \textit{text-to-motion} retrieval as an alternative to the \textit{query-by-example} search, and inherently different from the searching using a query label from a fixed pool of labels. 
We employed two state-of-the-art text-encoder networks, as well as widely adopted motion-encoder networks, for learning a common space and producing the first baselines for this novel task. 
We demonstrated that the CLIP text encoder works best also for encoding domain-specific natural sentences inherently different from image-descriptive ones, and that Transformers and GCNs obtain better motion representation than GRU-based encoders. 
In future works, we plan to train the models jointly on the two datasets and perform some cross-dataset evaluation to measure their generalization abilities and robustness. 
Other improvements include the use of video modality other than the motion and some unsupervised pre-training methods for boosting performance.

%



%
\begin{acks}
This research was supported by ERDF ``CyberSecurity, CyberCrime and Critical Information Infrastructures Center of Excellence'' (No. CZ.02.1.01/0.0/0.0/16\_019/0000822), by AI4Media -- A European Excellence Centre for Media, Society, and Democracy (EC, H2020 No. 951911), and by SUN -- Social and hUman ceNtered XR (EC, Horizon Europe No. 101092612).
\end{acks}

%

\bibliographystyle{ACM-Reference-Format}
\bibliography{references-Jan-disa,references-Jan,references-Nicola}

\end{document}